\documentclass[12pt, letterpaper]{article} 
\usepackage[margin=1in]{geometry}  

\usepackage{authblk}
\usepackage{amssymb}
\usepackage{amsmath}
\usepackage{amsmath,amssymb,amsfonts}
\usepackage{algorithmic}
\usepackage{multirow}
\usepackage{graphicx}
\newcommand{\cmark}{\ding{51}}%
\newcommand{\xmark}{\ding{55}}%
\usepackage{array}
\usepackage{pifont}

\def\bA{\mathbf{A}}

\def\bH{\mathbf{H}}
\def\bI{\mathbf{I}}

\def\bW{\mathbf{W}}
\def\bX{\mathbf{X}}

\def\bSigma{{\boldsymbol\Sigma}}

\def\bb{\mathbf{b}}

\def\be{\mathbf{e}}

\def\bz{\mathbf{z}}

\def\btheta{{\boldsymbol\theta}}

\def\bmu{{\boldsymbol\mu}}

\def\bpi{{\boldsymbol\pi}}

\def\bsigma{{\boldsymbol\sigma}}

\def\bphi{{\boldsymbol\phi}}

\def\bbE{\mathbb{E}}

\def\bbR{\mathbb{R}}

\def\cD{\mathcal{D}}

\def\cL{\mathcal{L}}

\def\cN{\mathcal{N}}

\def\diag{{\sf diag}}

\def\KL{{\sf KL}}

\def\zeros{\mathbf{0}}

\usepackage{textcomp}
\usepackage{xcolor}
\usepackage{comment}
\usepackage{booktabs}
\usepackage[colorlinks=true,
            linkcolor=blue,
            citecolor=blue,
            urlcolor=blue]{hyperref}

\title{Gait-Based Hand Load Estimation via Deep Latent Variable Models with Auxiliary Information}
\date{}

\author[1]{Jingyi Gao}
\author[2]{Sol Lim}
\author[1]{Seokhyun Chung\thanks{corresponding author; schung@virginia.edu}}
\affil[1]{Department of Systems and Information Engineering, University of Virginia}
\affil[2]{Department of Industrial and Systems Engineering, Virginia Polytechnic Institute and State University}

\begin{document}
\maketitle

\begin{abstract}

Machine learning methods are increasingly applied to ergonomic risk assessment in manual material handling, particularly for estimating carried load from gait motion data collected from wearable sensors. However, existing approaches often rely on direct mappings from loaded gait to hand load, limiting generalization and predictive accuracy. In this study, we propose an enhanced load estimation framework that incorporates auxiliary information, including baseline gait patterns during unloaded walking and carrying style. While baseline gait can be automatically captured by wearable sensors and is thus readily available at inference time, carrying style typically requires manual labeling and is often unavailable during deployment. Our model integrates deep latent variable modeling with temporal convolutional networks and bi-directional cross-attention to capture gait dynamics and fuse loaded and unloaded gait patterns. Guided by domain knowledge, the model is designed to estimate load magnitude conditioned on carrying style, while eliminating the need for carrying style labels at inference time. Experiments using real-world data collected from inertial measurement units attached to participants demonstrate substantial accuracy gains from incorporating auxiliary information and highlight the importance of explicit fusion mechanisms over naive feature concatenation.

\end{abstract}

\section{Introduction}

Work-related musculoskeletal disorders (WMSDs) are a major occupational health concern, especially in industries involving frequent manual material handling, such as manufacturing \cite{a1} and construction \cite{a2}. Prolonged and repeated heavy lifting increases the risk of chronic pain and fatigue, leading to lower productivity and higher healthcare costs \cite{a3}. As a result, accurately assessing and monitoring ergonomic exposure plays a key role in evaluating the ergonomic risk associated with WMSDs.

With recent advances in sensor technologies enabling continuous monitoring of workers’ movements, data-driven approaches are increasingly employed to enhance WMSD risk evaluation frameworks. Among these, machine learning (ML) has emerged as a particularly effective tool for estimating hand load during diverse manual material handling tasks, including load carriage and lifting, using motion data captured by wearable sensors \cite{a4,a5}. In this framework, ML models can learn complex relationships between motion kinematics and the magnitudes of loads being carried or lifted. Once trained, these models can generalize to new, unseen workers, providing real-time load predictions based on their motion data. These load estimates can subsequently be integrated into ergonomic risk models to quantify biomechanical strain, facilitate systematic evaluations of ergonomic risk factors, and inform the development of targeted injury prevention strategies.

Beyond gait and motion kinematics captured during load carriage, incorporating auxiliary factors can substantially enhance the predictive performance of ML models. In particular, we leverage two key features: (i) baseline gait patterns observed without carrying any objects and (ii) carrying style (e.g., whether the object is held with one or both hands) to improve load prediction. Baseline movement characteristics, such as walking speed, stride length, and cadence, can inherently reflect a worker's unique physical traits, which are critical for accurate load estimation. For example, a worker with a naturally slower gait may exhibit different load-bearing mechanics compared to someone with a faster pace, even if both carry the same weight \cite{a6,a7}. Similarly, carrying style adds important context to how the load interacts with the worker's body \cite{a8,a9,a10}. Incorporating these auxiliary features can therefore significantly improve the accuracy of load predictions.

A straightforward approach to incorporating auxiliary information would be to include it directly as part of the input to a prediction model. However, naively integrating such data introduces critical challenges. For instance, baseline gait motion patterns recorded without loads are often redundant, as they typically consist of repeated walking cycles with minimal variation. Without appropriate processing, this redundancy can obscure meaningful load-related features and ultimately dilute the model’s predictive performance. Moreover, gait kinematics are inherently temporal, with key biomechanical cues appearing only in specific phases of the gait cycle. Treating all time steps equally may cause the model to overlook brief but informative segments. Therefore, an effective predictive framework must include mechanisms that not only identify and emphasize relevant patterns in baseline gait data, but also discern and attend to the most informative temporal windows within the time-series signals.

Moreover, carrying style information is often unavailable at deployment. While gait patterns (both with and without loads) can be continuously captured in real time via wearable sensors, identifying the carrying style (e.g., whether the object is held with one or both hands) typically requires manual labeling, which is impractical at scale. As such, the absence of carrying style annotations during inference limits its direct inclusion in the model input, unless the model is specifically designed to accommodate such auxiliary data available only during training.

To this end, we propose a predictive framework based on a deep latent-variable model, termed \texttt{AuxVAE}. Built upon the variational autoencoder (VAE) architecture, \texttt{AuxVAE} learns to extract probabilistic latent representations from high-dimensional gait kinematics and leverages these representations to predict load magnitude via an integrated neural network regressor. This architecture enables simultaneous inference of latent variables (i.e., low-dimensional representations of gait patterns) and target outputs (i.e., load magnitudes). Here, a key strength of our approach lies in its ability to incorporate useful auxiliary information beyond gait motions during load carriage to enhance prediction accuracy.

The central mechanism for incorporating auxiliary unloaded gaits is the estimation of \textit{conditional} distributions for both the latent representations and the target variable. Specifically, the model infers the distribution of latent variables conditioned on unloaded gait, enabling the latent space to capture individual-specific kinematic characteristics. This allows the model to enrich its latent representations using individual gait patterns. To further enhance representation learning, the encoder is equipped with bidirectional cross-attention layers that selectively attend to informative temporal segments across both loaded and unloaded gait sequences.

Additionally, our model incorporates carrying style, another auxiliary variable available only during training. To achieve this, we design the neural network regressor to learn both the distribution over carrying modes and the distribution over load magnitudes conditioned on the carrying modes. During training, the model aims to maximize the marginal likelihood, integrating over the carrying style distribution. This enables the model to simultaneously learn to infer carrying style from latent representations and to estimate load magnitudes based on the inferred style. Although the true carrying style is unavailable at inference time, the model uses its inferred carrying style to condition the load estimate. Thus, this joint modeling framework allows the model to account for the influence of carrying style and benefit from its structure, even when it is not directly observable on the fly.

The contributions of this study can be summarized as follows.

\begin{itemize}
    \item We build a predictive framework for estimating \textit{hand load magnitudes from motion data}, designed to incorporate auxiliary information within a probabilistic deep generative modeling paradigm.
    \item We introduce an attention mechanism that personalizes the prediction model by effectively integrating baseline gait patterns, which \textit{reflect underlying individual-specific kinematic characteristics}.
    \item We propose a method that integrates carrying style information, an auxiliary variable available \textit{only during training}, to further enhance model performance. 
    \item We validate the proposed approach using a real-world gait kinematics dataset collected from inertial sensors worn by 22 healthy participants of varying ages while carrying boxes.
\end{itemize}

The remainder of the paper is organized as follows. In Section \ref{RelatedWork}, we review related literature. In Section \ref{Methodology}, we develop our proposed model. In Section \ref{Experiments}, we discuss experimental results. In Section \ref{Conclusion}, we conclude the paper and discuss the future work.

\section{Related Work} \label{RelatedWork} 

Early efforts on ergonomic risk assessment commonly relied on checklist-based methods such as the Rapid Upper Limb Assessment (RULA) \cite{r1} and the Rapid Entire Body Assessment (REBA) \cite{r2}. These classical approaches enable quick risk scoring through visual inspection, but they are often manual and labor-intensive, and are limited by observer subjectivity and low temporal resolution. Recent advancements in sensor and monitoring technologies have enabled the continuous collection of workers’ gait and movement data, creating new opportunities for ML and deep learning to address these shortcomings and serve as core components of modern ergonomic risk assessment frameworks. For instance, Jiao et al. \cite{r3} introduced a two-stage framework that reconstructs 3D human pose and subsequently applies a temporal convolutional neural network (CNN) to map action sequences to REBA scores. Similarly, computer vision methods have been developed to directly infer RULA scores from RGB video \cite{r4}, addressing the inherent limitations of manual scoring. 

In addition to predicting established risk scores, ML and deep learning have been actively leveraged in diverse tasks to promote occupational safety, such as posture recognition \cite{r5,r6} and biomechanical overload assessment \cite{r7}. In particular, the widespread availability of low-cost depth cameras and off-the-shelf pose estimation frameworks have enabled researchers to harness video data for ML-based ergonomic risk assessment. In these approaches, raw videos of workers performing tasks are transformed into skeleton-based pose sequences, which are then input into deep learning models to analyze movement patterns and assess physical risk levels. For instance, a variational deep learning architecture \cite{r8} was developed to estimate ergonomic risk using the REBA framework, where variational approaches process skeletal information to construct descriptive latent spaces for accurate human posture modeling. Zhou et al. \cite{r9} developed an attention-based graph CNN specifically designed to focus on hazardous frames across long video sequences for ergonomic risk assessment. 

However, vision-based systems are often impractical when occlusion is present or viewpoints are constrained. Wearable sensors offer a complementary solution by streaming motion data directly from the body. Antwi-Afari et al. \cite{r10} used a recurrent neural network (RNN)-based framework to effectively detect unsafe postures using plantar pressure data. Support vector machines (SVMs), gradient-boosted decision trees, and other ML classifiers have also been used to detect hazardous movements in real time based on distributed inertial measurement unit (IMU) inputs \cite{r11}. Furthermore, Li et al. \cite{r12} showed that the conditional VAE and conditional generative adversarial network (GAN) can effectively simulate worker postures during lifting tasks to facilitate the prevention of low back pain in occupational settings. These wearable sensor approaches have achieved remarkable accuracy, with some studies reporting over 99\% accuracy in distinguishing correct versus incorrect postures during lifting tasks \cite{r13}. 

While posture classification plays a critical role, injury prevention frameworks also rely on the continuous monitoring of physical loads imposed on the body during material handling tasks. For example, Matijevich et al. \cite{r14} identified that data from trunk-mounted IMU sensors and pressure insoles constitute key features for accurately monitoring low back loading, and employed a gradient-boosted decision tree algorithm to estimate lumbar moments. Notably, there have been efforts to estimate the exact load magnitude using wearable sensor data that capture motion during carrying tasks, as this information cannot be directly measured by the sensors themselves. For example, Lim and D’Souza \cite{r15} utilized gait kinematics to jointly predict the carrying mode and load level of object-handling tasks using random forests and a Bayesian framework. More recently, Rahman et al. \cite{r16} investigated potential algorithmic biases in ML-based hand load estimation, showing that models trained on sex-imbalanced populations may produce biased predictions and developing a VAE-based approach to mitigate such bias. 

Despite the promising results reported in prior studies, the existing literature in ergonomics remains largely centered on using ML and deep learning methods to infer target outcomes (e.g., hand loads or risk scores) based on aggregated data collected from workers (e.g., gait motions) engaged in ergonomically demanding tasks. In contrast, our approach advances beyond this framework by incorporating auxiliary information into the predictive framework for hand load estimation via deep latent variable modeling. This auxiliary information includes not only readily available auxiliary inputs (baseline gaits without loads) but also auxiliary targets (carrying style) that may not be directly observable at test time. 

Although not primarily aimed at ergonomic risk evaluation, recent efforts in gait analysis have focused on enhancing representation learning and predictive performance through multimodal data fusion; an objective shared by our study. In this domain, integrating heterogeneous data sources has proven effective in capturing complex biomechanical patterns. For instance, Cui et al. \cite{r17} introduced MMGaitFormer, a transformer-based framework that effectively fuses spatial-temporal information from skeletons and silhouettes. This approach employs spatial and temporal fusion modules to integrate complementary features, resulting in improved recognition performance under varying clothing conditions. Similarly, the Multi-Stage Adaptive Feature Fusion (MSAFF) neural network, proposed by Zou et al. \cite{r18}, performs multimodal fusion at different stages of feature extraction. By considering the semantic association between silhouettes and skeletons, MSAFF enhances the discriminative power of gait features. Moreover, Aung and Kusakunniran \cite{r19} provided a comprehensive review of gait analysis using deep learning approaches in criminal investigations, highlighting the importance of considering biometric factors such as gender, age, and outfits, which can influence gait dynamics.

Inspired by these advances, our framework leverages auxiliary information not simply as additional inputs but as structured signals to improve representation learning. This enables more accurate modeling of underlying data distributions and leads to superior predictive performance compared to approaches that overlook or naively incorporate auxiliary data.

\section{Proposed Model} \label{Methodology}

\subsection{Problem Description}

Suppose gait kinematics are recorded for $N$ workers using $S$ sensors attached to each worker, with data collected over a fixed number of gait cycles. Each worker performs $M$ load-carrying tasks, where each task is characterized by the load magnitude and one of $L$ predefined carrying styles. We denote the dataset as $\cD = \{\cD_{i,j}\}_{i=1,\,j=1}^{N,\, M} = \{(\bX_{i,j}, y_{i,j})\}_{i=1,\, j=1}^{N,\, M}$, where $i$ and $j$ index the workers and tasks, respectively. Here, $\mathbf{X}_{i,j} \in \mathbb{R}^{T \times S}$ represents the multivariate time-series signal from $S$ sensors recorded over $T$ time steps, and $y_{i,j} \in \mathbb{R}_+$ denotes the corresponding load magnitude. ML-based load magnitude estimation is a supervised learning task aimed to learn an underlying function $f: \mathbb{R}^{T \times S} \rightarrow \mathbb{R}_+ $ mapping the input gait kinematics $\mathbf{X}_{i,j}$ to the output load magnitude $y_{i,j}$. Once trained, the model is deployed to accurately infer the load magnitude for an unseen worker $i' \notin \{1, \ldots, N\}$ given a specific task.

In addition to $\mathcal{D}$, we have auxiliary data $\mathcal{D}^{\sf aux} = \{\mathcal{D}_{i,j}^{\sf aux}\}_{i=1,\,j=1}^{N,\,M} = \{ (\mathbf{X}^{\sf aux}_{i}, y^{\sf aux}_{i,j}) \}_{i=1,\, j=1}^{N,\, M}$, where $\mathbf{X}^{\sf aux}_{i} \in \mathbb{R}^{T_0 \times S}$ represents the gait kinematics of worker $i$ while walking without carrying a load, and $y^{\sf aux}_{i,j} \in \{0,1\}^L$ is a one-hot vector indicating the carrying style (from among $L$ possible styles) adopted by worker $i$ during task $j$. For a previously unseen worker $i'$, we assume that $\bX^{\sf aux}_{i'}$ is available at inference time, while $y^{\sf aux}_{i',j}$ is not. This assumption reflects that obtaining $y^{\sf aux}_{i',j}$ typically requires manual annotation and is impractical to obtain in real-world deployments. Given this setting, we refer to $\bX^{\sf aux}_{i}$ and $y^{\sf aux}_{i,j}$ as the auxiliary input and output, respectively. Our goal is to accurately estimate $f$ by learning a predictive model $f_{\boldsymbol{\theta}}$ parameterized by $\boldsymbol{\theta}$, using the primary dataset $\mathcal{D}$, with the help of auxiliary information from $\mathcal{D}^{\sf aux}$ to enhance model performance.

For clarity, we omit the indices $i$ and $j$ when referring to a single data instance, whenever the context allows.

\subsection{Model Development}

We now turn to the development of our model, which is grounded in deep latent variable modeling. At its core is the VAE framework, a generative model based on probabilistic autoencoding \cite{m1}. Although VAEs were originally introduced for generative tasks, their ability to model complex data distributions through latent variable representations makes them well-suited for supervised learning from high-dimensional data. Motivated by this capacity, we leverage VAEs to learn the distribution of gait kinematics, represented as high-dimensional time-series data collected from multiple wearable sensors. 

\begin{figure*}[htb!]
\centerline{\includegraphics[width=13cm]{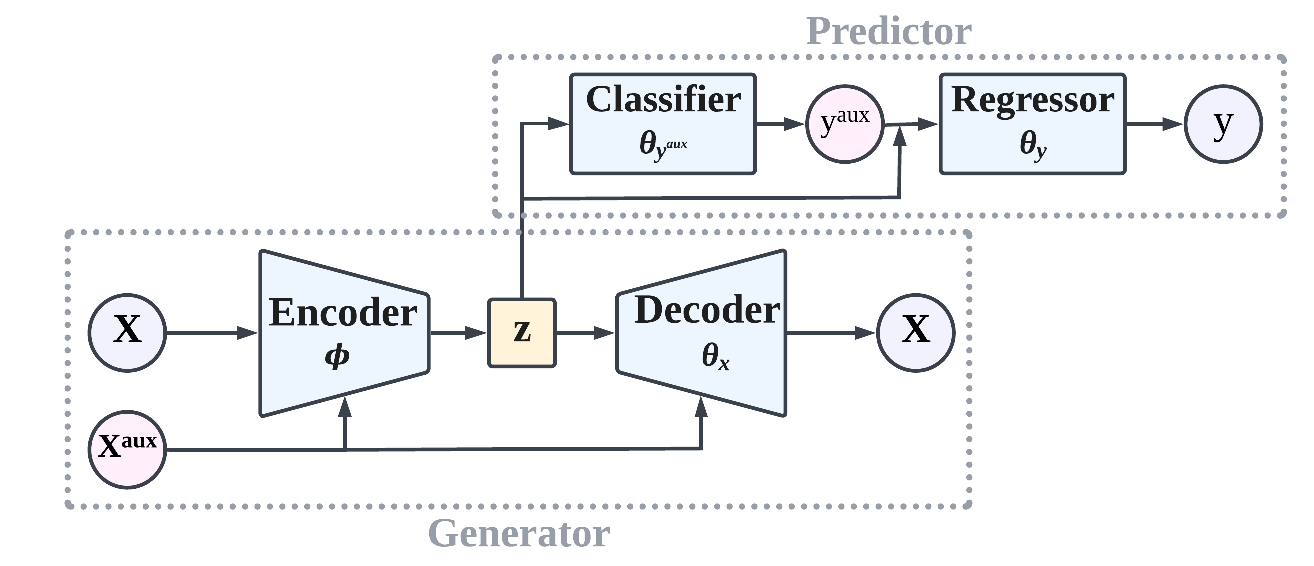}}
\caption{The structure of our proposed model.} 
\label{fig:model}
\end{figure*}

Before diving into the mathematical details, we first introduce the overall structure of our proposed model. Our proposed model, illustrated in Figure \ref{fig:model}, contains two main components:

\begin{itemize}
    \item A generative model that is fed and reconstructs loaded gait signals $\bX$ to extract latent representations $\bz$, conditioned on auxiliary signals $\bX^{\sf aux}$.
    \item A regression neural network that models the load magnitude $y$ conditioned on latent representation $\bz$ and the carrying style $y^{\sf aux}$, inferred with an auxiliary classification neural network predicting $y^{\sf aux}$ from $\bz$.
\end{itemize}

We next describe the development of our proposed model and its inference in detail.

\subsubsection{Building a deep latent variable model}

Building on the deep latent variable modeling framework \cite{m2}, we introduce a latent variable $\bz$ with which the joint distribution is defined as
\begin{equation}\label{eq:joint}
    p_\btheta(\bX, y, y^{\sf aux}, \bz \vert \bX^{\sf aux})
\end{equation}
over the latent variable $\bz$ and observed variables $\bX, y,$ and $y^{\sf aux}$, conditioned on auxiliary input $\bX^{\sf aux}$. The distribution \eqref{eq:joint} is parameterized by a deep neural network, with all associated parameters collectively denoted by $\btheta$.

We assume the following factorization of the joint distribution \eqref{eq:joint}:
\[
p_\btheta(\bX, y, y^{\sf aux}, \bz \vert \bX^{\sf aux}) = p(\bz)p_{\btheta_\bX}(\bX \vert \bX^{\sf aux}, \bz)p_{\btheta_y}(y \vert y^{\sf aux}, \bz)p_{\btheta_{y^{\sf aux}}}(y^{\sf aux} \vert \bz),
\]
where each component is specified as
\begin{align}
    p(\bz) &= \cN\left(\bz; \zeros,\bI\right), \\
    p_{\btheta_\bX}(\bX \vert \bX^{\sf aux}, \bz) &= \cN\left(\bX; \bmu_{\bX}, \diag(\bsigma_{\bX})\right),\label{eq:p(X)}\\
    p_{\btheta_{y^{\sf aux}}}(y^{\sf aux} \vert \bz) &= \text{Categorical}\left(y^{\sf aux} ; \bpi\right), \label{eq:p(y_aux)}\\
    p_{\btheta_y}(y \vert y^{\sf aux},\bz) &= \cN\left(y; \mu_{y}, \sigma_{y}\right)\label{eq:p(y)}.
\end{align} 
with $\btheta \equiv \{\btheta_\bX, \btheta_{y^{\sf aux}}, \btheta_y\}$ and $\diag(\cdot)$ being a diagonal matrix corresponding to a given vector; $\text{Categorical}\left(\cdot ; \bpi\right)$ being the probability mass function of the categorical distribution with 
$\bpi$ being the class probabilities; $\cN(\cdot; \bmu, \bSigma)$ being the probability density function of the Gaussian distribution with mean $\bmu$ and covariance $\bSigma$. For notational convenience, we use
$\bX$ to denote both the matrix form and its vectorized form when used as an argument in probability distributions.

The parameters of the above distributions \eqref{eq:p(X)}--\eqref{eq:p(y)} depend on deterministic transformations of their respective inputs, computed via neural networks as follows:
\begin{align}
    (\bmu_\bX, \bsigma_\bX) &= \textrm{NeuralNet}(\bX^{\sf aux}, \bz; \btheta_\bX),\label{eq:neur(X)}\\
    \bpi &= \textrm{NeuralNet}(\bz; \btheta_{y^{\sf aux}}), \label{eq:neur(y_aux)}\\
    (\mu_y, \sigma_y) &= \textrm{NeuralNet}(y^{\sf aux}, \bz; \btheta_{y})\label{eq:neur(y)}.  
\end{align}
Here, $\textrm{NeuralNet}(\cdot;\btheta)$ denotes the output of a neural network parameterized by $\btheta$, applied to the specified input. 

\subsubsection{Learning via variational Bayes}

Direct maximum likelihood estimation of the proposed deep latent variable model is generally infeasible. This is due to the intractability of the marginal likelihood $$p_\btheta(\bX, y, y^{\sf aux} \vert \bX^{\sf aux}) = \int p_\btheta(\bX, y, y^{\sf aux}, \bz \vert \bX^{\sf aux})d\bz,$$ which also renders the exact posterior $p_\btheta(\bz \vert \bX, y, y^{\sf aux}, \bX^{\sf aux})$ intractable. We adopt variational Bayes (or variational inference) to address this issue \cite{m1}. Variational Bayes seeks to approximate the intractable true posterior with a tractable surrogate distribution, known as the variational posterior distribution. This is typically achieved by minimizing the Kullback–Leibler (KL) divergence between the variational posterior and the exact posterior. This minimization is mathematically equivalent to maximizing the evidence lower bound (ELBO). Optimizing the ELBO circumvents the intractability of direct marginal likelihood computation.

Specifically, we introduce a variational posterior $q_\bphi(\bz \vert \bX, \bX^{\sf aux}) \approx p_\btheta(\bz \vert \bX, y, y^{\sf aux}, \bX^{\sf aux})$, specified as 
\begin{equation}\label{eq:q(z)}
    q_\bphi(\bz \vert \bX, \bX^{\sf aux}) = \cN(\bz; \bmu_\bz, \diag(\bsigma_\bz) )
\end{equation}
parameterized by a deep neural network with parameters $\bphi$:
\begin{equation}
    (\bmu_\bz, \bsigma_\bz) = \text{NeuralNet}(\bX, \bX^{\sf aux}; \bphi). \label{eq:neur(z)}
\end{equation}

Given \eqref{eq:q(z)}, we decompose the marginal log-likelihood to derive the ELBO:
\begin{align}
    &\log p_\btheta(\bX,y,y^{\sf aux}\vert\bX^{\sf aux})\nonumber \\ 
    &= \bbE_{q_\bphi(\bz|\bX,\bX^{\sf aux})}\left[\log\frac{p_\btheta(\bX,y,y^{\sf aux},\bz|\bX^{\sf aux})}{q_\bphi(\bz|\bX,\bX^{\sf aux})}\right]+{\sf KL}(q_\bphi(\bz|\bX,\bX^{\sf aux})\Vert p_\btheta(\bz|\bX,y,y^{\sf aux},\bX^{\sf aux})),\label{eq:KL}
\end{align}
where $\sf{KL}(\cdot \Vert \cdot)$ represents the KL divergence between two specified distributions. The first term on the right-hand side corresponds to the ELBO. Since the second term is always non-negative by the definition of the KL divergence, we obtain the following inequality: 
\begin{equation}
    \log p_\btheta(\bX,y,y^{\sf aux}\vert\bX^{\sf aux})  \ge \bbE_{q_\bphi(\bz|\bX,\bX^{\sf aux})}\left[\log\frac{p_\btheta(\bX,y,y^{\sf aux},\bz|\bX^{\sf aux})}{q_\bphi(\bz|\bX,\bX^{\sf aux})}\right] := \cL(\btheta, \bphi)
\end{equation}
demonstrating that the ELBO $\cL(\btheta, \bphi)$ serves as a lower bound on the marginal log-likelihood. This ELBO then can be reorganized as:
\begin{align} 
&\cL(\btheta,\bphi) =\bbE_{q_{\phi}(\bz\vert\bX,\bX^{\sf aux})}[\log p_{\btheta_{\bX}}(\bX\vert\bX^{\sf aux},\bz)+\log p_{\btheta_{y^{\sf aux}}} (y^{\sf aux}\vert\bz)+\log p_{\btheta_y} (y\vert y^{\sf aux},\bz)] \nonumber\\
&\qquad\qquad\> -\KL[q_{\bphi}(\bz\vert\bX,\bX^{\sf aux})\Vert p(\bz)].\label{eq:finalELBO}
\end{align}
Model training is done by maximizing $\cL(\btheta, \bphi)$ evaluated on the training samples, with respect to $\{\btheta, \bphi\}$.

The rearrangement in \eqref{eq:finalELBO} enables a clear interpretation of each term in the ELBO. The first term assesses how effectively the generative model with latent variable $\bz$ can explain the observed data $\bX, y, y^{\sf aux}$, given auxiliary information $\bX^{\sf aux}$. The second term penalizes divergence between the variational posterior $q_\bphi(\bz \vert \bX, \bX^{\sf aux})$ and the prior $p(\bz)$ specified as a standard Gaussian. This KL divergence acts as a regularizer, discouraging the learned posterior to stray too far from the assumed prior distribution $p(\bz)$. 

In particular, examining the individual components of the first term in \eqref{eq:finalELBO}:
\begin{itemize}
    \item Reconstruction term: 
    $$\bbE_{q_{\bphi}(\bz \vert \bX, \bX^{\sf aux})}[\log p_{\theta}(\bX\vert\bX^{\sf aux},\bz)]$$ encourages the model to generate data that closely matches the observed input $\bX$ when sampling $\bz$ from the variational posterior that is conditioning on $\bX^{\sf aux}$. Together with the KL term $\KL[q_{\bphi}(\bz\vert\bX,\bX^{\sf aux})\Vert p(\bz)]$, this component facilitates the learning of well-regularized latent representations of $\bX$ that effectively incorporate information from the auxiliary input $\bX^{\sf aux}$. These learned latent representations are leveraged for prediction of the target variable $y$. 
    
    \item Supervised learning terms: $$\bbE_{q_{\bphi}(\bz \vert \bX, \bX^{\sf aux})}[p_\btheta(y^{\sf aux}\vert\bz)]  \textrm{ and } \bbE_{q_{\bphi}(\bz \vert \bX, \bX^{\sf aux})}[p_\btheta(y\vert y^{\sf aux},\bz)]$$ enable the joint modeling of carrying style and load magnitude. This multi-task design leverages the auxiliary label on carrying style $y^{\sf aux}$ by conditioning the prediction of load magnitude $y$ on it. Such a structure integrates domain knowledge, recognizing that the estimation of hand load from gait motion is highly dependent on how the load is carried \cite{r15}.
\end{itemize}

Based on our proposed deep latent variable model and its corresponding approximate posterior, we construct a unified neural network architecture, as depicted in Figure~\ref{fig:model}. The encoder $\text{NeuralNet}(\bX, \bX^{\sf aux}; \bphi)$ takes as input both $\bX$ and auxiliary features $\bX^{\sf aux}$ to parameterize the approximate posterior distribution $q_\bphi(\bz \vert \bX, \bX^{\sf aux})$. The resulting latent representation $\bz$ is then decoded back into the original feature space via the decoder $\text{NeuralNet}(\bz, \bX^{\sf aux}; \btheta_\bX)$, which models $p_{\btheta_\bX}(\bX \vert \bz, \bX^{\sf aux})$. Together, the encoder and decoder constitute a conditional VAE, labeled as the ``Generator" in Figure~\ref{fig:model}. Concurrently, the latent variable $\bz$ is utilized to construct the predictive distribution $p_{\btheta_y}(y \vert y^{\sf aux}, \bz)$ for the target variable $y$. This prediction is performed jointly by a classifier $\text{NeuralNet}(\bz; \btheta_\bz)$ and a regressor $\text{NeuralNet}(y^{\sf aux}, \bz; \btheta_{y^{\sf aux}})$, collectively referred to as the ``Predictor" in Figure~\ref{fig:model}.

\subsection{Enhanced fusion of temporal auxiliary inputs}
\begin{figure}[htb!]
\centerline{\includegraphics[width=17cm]{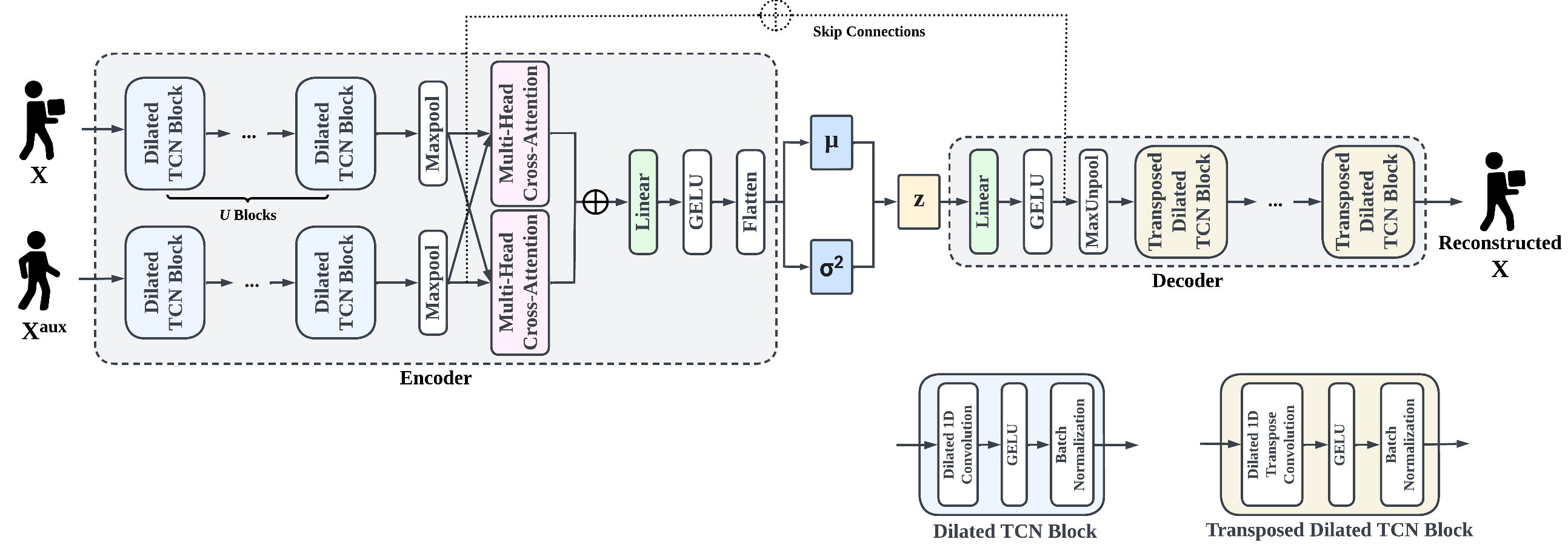}}
\caption{The ``Generator'' equipped with TCNs and bi-directional cross-attention.}
\label{fig:encoder}
\end{figure}

Our model integrates the auxiliary input $\bX^{\sf aux}$ into both the encoding and decoding processes of the primary input $\bX$ to improve the performance of downstream prediction tasks. Since both $\bX^{\sf aux}$ and $\bX$ are temporal and multi-stream in nature, it is critical to extract informative temporal features from each stream and selectively attend to the relevant segments of each signal. To achieve this, we employ a dual-stream encoder composed of dilated temporal convolutional networks (TCNs) \cite{m3} and a bidirectional cross-attention mechanism \cite{m4}. The architecture of our encoder and following decoder is illustrated in Figure \ref{fig:encoder}. 

\subsubsection{Temporal feature extraction via dilated TCNs} \label{sec:tcn}

We employ TCN blocks in the autoencoder architecture. The encoder processes both $\bX$ and $\bX^{\sf aux}$ using stacked 1D TCN blocks, each with progressively increasing receptive fields. Similarly, the decoder consists of stacked TCN blocks to reconstruct $\bX$. Each TCN block begins with a dilated temporal convolution layer in the encoder, and a transposed dilated temporal convolution layer in the decoder. These are followed by a GELU activation function \cite{m5} and batch normalization, applied within each block for enhanced learning and training stability.

To illustrate the mechanism of dilated temporal convolution, let $\bH^{(u-1)}$ denote the input to the $u$-th TCN layer, initialized with $\bH^{(0)} = \bX$. Specifically, the entry $\bH^{(0)}_{t,s}$ represents the observation from sensor $s$ at time $t$ (i.e., $\bX_{t,s}$).
Given that layer $u$ has $S^{(u)}$ channels, the dilated convolution operation is expressed as
\begin{equation}
    \bH^{(u)}_{t,\>s} = \sum_{s'=1}^{S^{(u)}}\sum_{k=0}^{K-1}\bW^{(u)}_{k, \> s,\>s'}\cdot \bH^{(u-1)}_{t-d^{(u)}k,\> s'}+\bb^{(u)}_{s} \nonumber
\end{equation}
where $\bW^{(u)}_{k, \> s,\>s'}$ is the $k$-th convolution kernel weight connecting input channel $s$ to output channel $s'$; and $\bb^{(u)}_s$ is a bias term. 

\begin{figure}[htb!]
\centerline{\includegraphics[width=8cm]{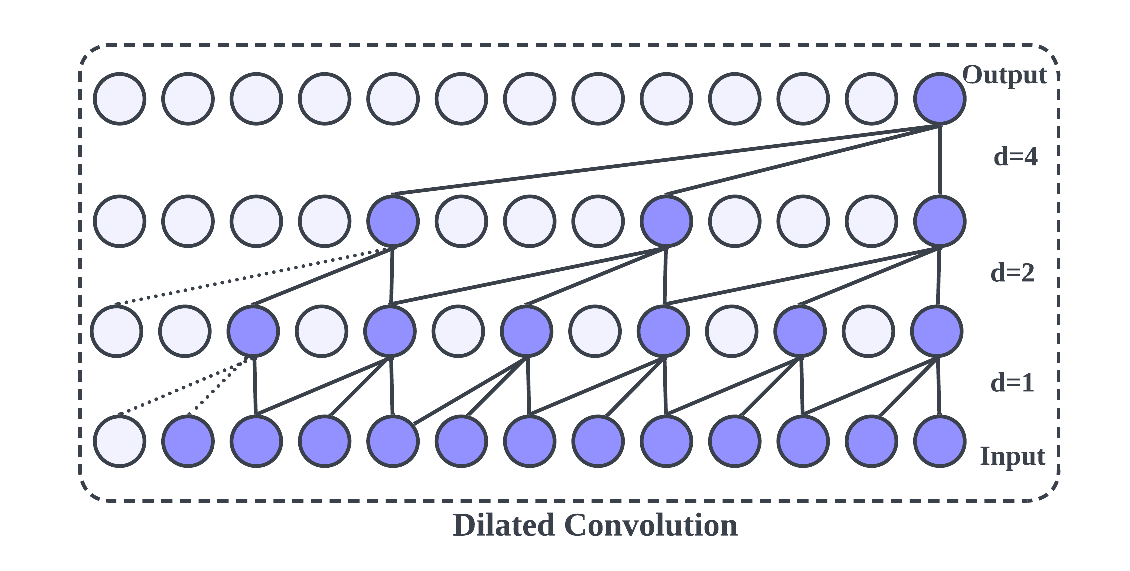}}
\caption{An example of dilated convolutional layers.}
\label{fig:dilated}
\end{figure}

This formulation enables each entry in a layer to aggregate input features from the previous layer over a dilated temporal window of size $k$, spaced by dilation factor $d$ (see Figure~\ref{fig:dilated}). We define the dilation rate $d^{(u)}=2^{u-1}$ to be exponentially growing over layers. A schematic illustration of dilated convolution is presented at the bottom right of Figure~\ref{fig:encoder}.  The same operation is applied to $\bX^{\sf aux}$. A transposed temporal convolution layer inverts the temporal downsampling process above, enabling the reconstruction of the original sequence resolution.

The dilated convolutional layers offer two key advantages. First, they impose a causal structure in which the representation at time step $t$ in a given layer depends only on information from time steps $\leq t$ in the preceding layer. This design ensures that temporal causality is preserved by preventing information leakage from the future. Second, dilation effectively expands the receptive field across layers without incurring significant computational overhead. This enables the model to capture long-range temporal dependencies without requiring excessive depth or computational cost.

\subsubsection{Bi-directional cross-attention}

When jointly encoding $\bX$ and $\bX^{\sf aux}$, simple concatenation of the input sequences is insufficient for identifying and emphasizing regions within the signals that are salient for prediction. We hypothesize that enabling the model to attend to informative segments across both the primary and auxiliary inputs, particularly those relevant to hand-load estimation, can enhance the quality of latent representations and improve predictive performance. To achieve this, we equip the encoder with bidirectional multi-head cross-attention.

Let $\bH \in \bbR^{T'\times d_h}$ and $\bH^{\sf aux} \in \bbR^{T'_0 \times d_h}$ denote the encoded representations of $\bX$ and $\bX^{\sf aux}$, respectively, obtained via TCN layers and max-pooling described in Section \ref{sec:tcn}. A set of $P$ attention heads is used to build bidirectional cross-attention between $\bH$ and $\bH^{\sf aux}$. For each head $p=1,\cdots,P$, we compute attention score matrices $\bA^{(p)} \in \bbR^{T' \times T'_0}$ and $\bA^{{\sf aux}(p)} \in \bbR^{T'_0 \times T'}$, where each entry quantifies the degree of attention between segments of $\bH$ and $\bH^{\sf aux}$. These attention weights are computed using the scaled dot-product \cite{m6}, as follows:
\begin{align*} 
&\bA^{(p)} = \textrm{softmax}\left(\frac{(\bH \bW_Q^{(p)})(\bH^{\sf aux} \bW_K^{(p)})^\top}{\sqrt{d_k}}\right), \\ 
&\bA^{{\sf aux}(p)} = \textrm{softmax}\left(\frac{(\bH^{\sf aux} \bW_Q^{{\sf aux}(p)})(\bH \bW_K^{{\sf aux}(p)})^\top}{\sqrt{d_k}}\right),
\end{align*}
where $\textrm{softmax}(\cdot)$ computes the row-wise softmax operation, and $\bW^{(p)}_Q, \bW^{(p)}_K\in \bbR^{d_h \times d_k}$ are the learnable query and key projection matrices for head $p$. The scalar $d_k$ denotes the dimensionality of the projected keys. The $(m,n)$-th entry of $\bA^{(p)}$ represents the attention score from the $m$-th temporal position in $\bH$ to the $n$-th position in $\bH^{\sf aux}$, indicating the extent to which the model attends to the auxiliary sequence at time $n$ when processing the primary sequence at time $m$. Similarly, the $(n,m)$-th entry of $\bA^{{\sf aux}(p)}$ captures the attention from $\bH^{\sf aux}$ to $\bH$.

After computing the attention scores, we derive the cross-attended representations as follows:
\begin{align} &\tilde{\bH}^{(p)} = \bA^{(p)}\bH^{\sf aux} \bW_V^{{\sf aux}(p)}\in\bbR^{T'\times d_v}, \nonumber\\
&\tilde{\bH}^{{\sf aux}(p)} = \bA^{{\sf aux}(p)} \bH \bW_V^{(p)}\in\bbR^{T_0'\times d_v}, \nonumber
\end{align}
where $\bW_V^{(p)}, \bW_V^{{\sf aux}(p)} \in \bbR^{d_h \times d_v}$ are the learnable value projection matrices for head $p$, and $d_v$ denotes the dimensionality of the projected value vectors. The matrices $\tilde{\bH}
^{(p)}$ and $\tilde{\bH}^{{\sf aux}(p)}$ represent the contextualized features of $\bH$ and $\bH^{\sf aux}$, respectively, after attending to the other input.

Then, we concatenate all attention heads:
\begin{align}
&\tilde{\bH} = \textrm{concat}\left(\tilde{\bH}^{(1)},\cdots,\tilde{\bH}^{(P)}\right)\bW_O \in\bbR^{T'\times(P\cdot d_v)} \nonumber\\
&\tilde{\bH}^{\sf aux} = \textrm{concat}\left(\tilde{\bH}^{{\sf aux}(1)},\cdots, \tilde{\bH}^{{\sf aux}(P)}\right)\bW^{\sf aux}_O \in\bbR^{T_0'\times(P\cdot d_v)} \nonumber
\end{align}
where $\bW_O,\bW^{\sf aux}_O \in \bbR^{P\cdot d_v\times P\cdot d_v}$ also are the learnable linear transformation matrices; and $\textrm{concat}(\cdot)$ concatenates matrices. Finally, the attention outputs for the loaded and unloaded sequences are concatenated as:
\begin{equation}
\tilde{\bH}^{\sf fused} = \textrm{concat}\left(\tilde{\bH},\tilde{\bH}^{\sf aux}\right)\in \bbR^{(T'+T'_0)\times (P\cdot d_v)}, \nonumber
\end{equation}
which is fed to the next layer.

Intuitively, by attending jointly to $\bH$ and $\bH^{\sf aux}$, the cross-attention mechanism may learn to isolate temporal segments where carrying a load induces the greatest deviations from a subject's baseline walking pattern. In effect, each attention head may focus on comparing corresponding phases of the gait cycle (e.g., heel strike, mid-stance, toe-off) between the two sequences, highlighting subtle shifts in timing or amplitude that correlate most strongly with the load magnitude $y$. That said, the cross-attention mechanism does not constrain the model to consider only corresponding gait phases. For example, it may discover that a heel strike segment in the unloaded gait is most informative when compared to a mid-stance segment in the loaded gait. Moreover, it does not require comparisons to be made over coarse segment blocks. Instead, it operates at a temporal resolution across $T$ and $T_0$ time points, enabling the model to learn granular time point-wise associations across the two sequences. This flexibility allows the attention mechanism to capture more nuanced and potentially non-obvious relationships between the inputs. Furthermore, the bidirectional design of our cross-attention enables information to flow both ways: not only does the loaded gait inform which parts of the unloaded gait are most relevant, but the unloaded gait also guides the model toward the salient regions of the loaded sequence. Unlike unidirectional cross-attention, which contextualizes only one input based on the other, our approach allows for mutual contextualization of both sequences. This reciprocal attentional flow encourages the encoder to learn a richer latent representation that explicitly captures how load affects gait patterns, ultimately improving the accuracy and robustness of the hand-load prediction.

\subsection{Implementation Specifics}

This section discusses some details and considerations for practitioners that implement our approach in practice.

\paragraph{The loss and KL annealing} In practice, we use a modified ELBO with a scaling factor $\beta$ on the KL term \cite{m7}, expressed as
\begin{align} 
\cL_\beta(\btheta,\bphi) = &\> \bbE_{q_{\bphi}(\bz\vert\bX,\bX^{\sf aux})}[\log p_{\btheta_\bX}(\bX\vert\bX^{\sf aux},\bz)+\log p_{\btheta_{y^{\sf aux}}}(y^{\sf aux}\vert\bz)+\log p_{\btheta_y}(y\vert y^{\sf aux},\bz)] \\ &
-\beta\KL[q_{\bphi}(\bz\vert\bX,\bX^{\sf aux})\Vert p(\bz)]. \nonumber
\end{align} We set $\beta$ to gradually increase from 0 to 1 over the course of training epochs, allowing the model to balance reconstruction fidelity with regularization of the approximate posterior distribution. This scheduling strategy helps mitigate posterior collapse, a common issue in VAEs where the latent variables easily become uninformative \cite{m8}. Furthermore, to evaluate the predictions $\hat \bX$, $\hat y^{\sf aux}$, and $\hat y$, we compute the mean squared error (MSE), cross-entropy, and mean absolute error (MAE), respectively. These losses correspond to assessing the respective conditional likelihoods $p_{\btheta_\bX}(\bX \vert \bX^{\sf aux}, \bz)$, $p_{\btheta_{y^{\sf aux}}}(y^{\sf aux} \vert \bz)$, and $p_{\btheta_y}(y \vert y^{\sf aux}, \bz)$. Note that, in place of MSE, we use MAE to evaluate $y$ as we empirically found that it yielded more stable and faster convergence during training in our case.

\paragraph{Training} Training our proposed model is done by solving an optimization problem: $$\min_{\btheta, \bphi} - \frac{1}{NM}\sum_{i=1;j=1}^{N;M}\cL_\beta(\btheta, \bphi; \cD_{i,j}, \cD^{\sf aux}_{i,j}),$$ which minimizes an expected loss over training samples collected from $N$ training participants for $M$ tasks. For clarity, we here let $\cL_\beta(\btheta, \bphi; \cdot)$ explicitly express the lowerbound evaluated on a specific training sample. We use the reparametrization trick \cite{m1} to enable low-variance stochastic gradient estimation, allowing efficient gradient-based optimization via backpropagation. 

\paragraph{Prediction} A key premise of this study is that the ground-truth carrying style $y^{\sf aux}$ (the auxiliary output) is not available for unseen workers. Crucially, our method does not require access to $y^{\sf aux}$ when predicting the load magnitude $y$. Given the primary and auxiliary gaits $\bX_{i'}$ and $\bar\bX_{i'}^{\sf aux}$ for a previously unseen worker $i'$, the prediction is computed as
\begin{equation}
\hat{y}=\frac{1}{S}\sum^S_{s=1}\sum^L_{l=1}\hat\pi^{(l,s)}\hat \mu^{(l,s)}_y,
\end{equation}
where $\hat{\mu}_y^{(l,s)}$ is the output of $\text{NeuralNet}(\be^{(l)}, \bz^{(s)}; \btheta_y)$ in \eqref{eq:neur(y)}, with $\be^{(l)}$ denoting the one-hot vector of carrying style $l$, and $\bz^{(s)}$ an \textit{i.i.d.} sample from the distribution established with $\text{NeuralNet}(\bX_{i'}, \bX^{\sf aux}_{i'}; \bphi)$ in \eqref{eq:neur(z)}. The corresponding weight $\hat{\pi}^{(l,s)}$ is the predicted probability of carrying style $l$, computed from $\text{NeuralNet}(\bz^{(s)}; \btheta_{y^{\sf aux}})$ in \eqref{eq:neur(y_aux)}. This formulation enables marginalization over possible $L$ carrying styles, allowing accurate prediction of $y$ without requiring explicit knowledge of $y^{\sf aux}$ at inference time.

\section{Experiments} \label{Experiments}
In this section, we present a comprehensive evaluation of our proposed \texttt{AuxVAE}. Our experiments are designed to appraise the model in three key aspects:
\begin{itemize}
    \item We compare the model’s overall prediction accuracy against several state-of-the-art baselines.
    \item We analyze the contribution of the cross-attention mechanism to feature fusion, contrasting it with simpler alternatives such as direct concatenation.
    \item We evaluate the benefits of jointly modeling auxiliary outputs, demonstrating how this strategy enhances the main prediction task.
\end{itemize}

\subsection{Setup}

\subsubsection{Dataset: IMU-based gait motions with hand load} 
\begin{figure}[htb!]
\centerline{\includegraphics[width=\textwidth]{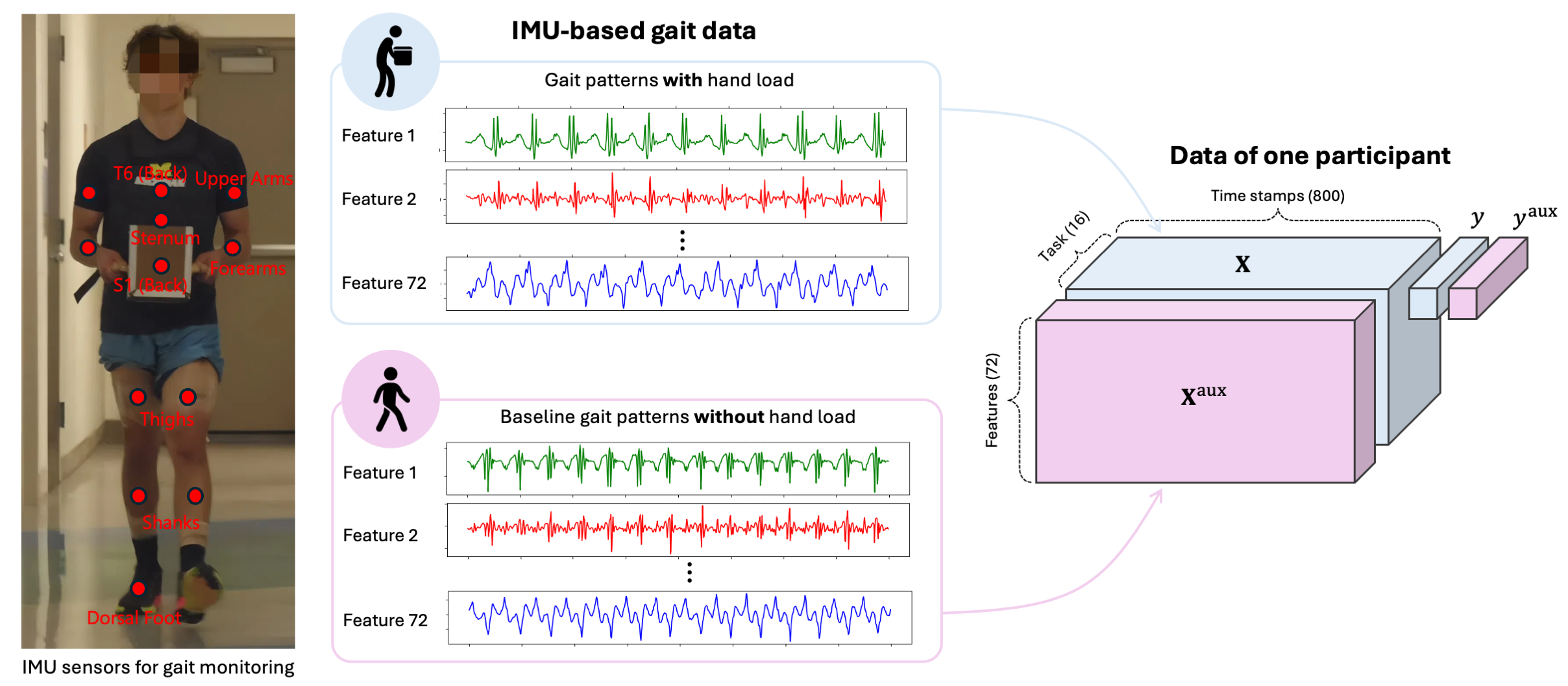}}
\caption{Illustration of the data structure for a single participant. A total of 22 participants contributed the entire dataset. Each task is defined by a unique combination of box weight ($y$) and carrying style ($y^{\sf aux}$), yielding 16 distinct tasks (4 weights $\times$ 4 styles). Note that although data collection was conducted at four discrete box weights, the target variable $y$ (box weight) is treated as a continuous variable. For each task, 72 sensor features (12 IMU sensors $\times$ 6 motion-related measurements) were recorded over 800 time steps.}
\label{fig:data}
\end{figure}

A dataset recording IMU-based gait motions with hand loads collected by \cite{r15} is used in this study. Figure \ref{fig:data} illustrates the data structure for a single participant. A total of 22 healthy participants of varying ages (12 males, 10 females; aged 18-55 years, mean 34.4 $\pm$ 10.9 years) took part in the study. Their average height was 68.4 $\pm$ 3.1 inches, weight 167.8 $\pm$ 29.5 pounds, and BMI 25.1 $\pm 3.4 \text{lb/in}^2$. All participants were screened for back injuries or chronic pain prior to the experiment and confirmed to have no such issues. Participants carried a box along a 24-meter corridor under four carrying styles: one-handed right, one-handed left, two-handed side, and two-handed anterior. Each mode was tested with four load magnitudes (a box of 10, 20, 30 and 50 pounds). Therefore, there are a total of 16 loaded conditions for each participant (4 carrying styles $\times$ 4 load levels), which were presented in a random order for each participant. Additionally, each participant completed a trial walking without carrying a box to record baseline gait patterns without any hand load. To capture natural adaptations to varying load conditions, participants were allowed to walk at a self-selected pace. A two-minute rest period was provided between trials to reduce fatigue and its potential influence on gait.

A set of IMU sensors attached to the participants recorded gait data for each walk. Twelve commercial sensors (Biostamp RC, mc10 Inc., Lexington, MA, USA) were placed at specific anatomical locations: the left and right thighs, left and right shanks, right dorsal foot, left and right upper arms, left and right forearms, sixth thoracic vertebra (T6), sternum, and first sacral vertebra (L5/S1). The sensors on the left and right shanks were primarily used to identify key gait events. Each IMU recorded tri-axial acceleration and angular velocity, yielding six motion attributes per sensor location at a sampling frequency of 80 Hz. As such, gait data for each trial is characterized by 72 motion attributes (6 motion attributes $\times$ 12 sensors). To ensure consistency across trials, we performed resampling of the time-series observations for each trial to a fixed length, enabling standardized input dimensions for downstream numerical studies.

\subsubsection{Model configurations}

All deep learning models were implemented in \textit{Pytorch (version 2.4.1)}. In particular, the encoder of \texttt{AuxVAE} consists of two dilated TCN layers with increasing dilation rates $\{1,2\}$, and the number of hidden units are $\{256, 128, 64\}$, respectively. The cross-attention mechanism employs 4 attention heads. The latent space dimension $k$ is set to 128. Models were trained using the Adam optimizer with an initial learning rate of $1\times10^{-3}$ and a batch size of 128, for a maximum of 500 epochs. The learning rate decreases by a factor of 0.1 every 100 epochs. Regularization was applied with a weight decay of $1\times10^{-4}$. A leave-one-participant-out cross-validation strategy was employed for training. Each experiment was repeated 10 times for evaluation.

\subsection{Experimental results}
\label{sec:exp_results}

\subsubsection{Experiment I: Comparison with benchmark models}

In this experiment, we evaluate the predictive performance of \texttt{AuxVAE} against several widely used benchmark models for time-series analysis. These include an RNN-based model (LSTM \cite{e1}), a CNN-based model (TCN \cite{m3}), and transformer-based models (Transformer \cite{m6}, Informer \cite{e2}, and TimesNet \cite{e3}). At the time of our study, TimesNet was generally recognized as a state-of-the-art approach for modeling time series data \cite{e4}. While these models were originally designed for general time-series prediction, we made minor modifications to adapt them for hand load estimation conditioned on carrying styles. All benchmarks receive both primary and auxiliary inputs to ensure a fair comparison. Since the benchmark models lack mechanisms for explicit fusion of these inputs, we provide them as concatenated vectors.

Table \ref{tab:baseline_comparison} presents the results, including average classification accuracies for carrying style and mean absolute errors in box weight estimation, computed over multiple runs with different random seeds. These results yield several key insights. First, \texttt{AuxVAE} consistently outperforms all benchmark models in both classification accuracy and load estimation error, highlighting the effectiveness of our proposed approach—specifically, the auxiliary fusion via bi-directional cross-attention and joint modeling. Notably, the poor load estimation performance of LSTM suggests that inaccurate predictions of carrying mode can significantly degrade hand load estimation. This underscores the benefit of our joint modeling framework. Another important observation is that TimesNet, despite being widely regarded as a state-of-the-art model for general time-series data, underperforms in our setting. This is the case even though it receives both primary and auxiliary inputs. This result suggests that simply concatenating auxiliary and primary data and feeding them into off-the-shelf models is insufficient. Instead, an explicit fusion mechanism, such as our cross-attention module, is crucial for effectively leveraging auxiliary signals.

\begin{table}[htb!]
\centering
\caption{Comparison with baseline models under auxiliary fusion settings.}
\label{tab:baseline_comparison}
\resizebox{\textwidth}{!}{%
\begin{tabular}{@{}ccccc@{}}
\toprule
Model              & $\bX^{\sf aux}$ Fusion Type & If Joint Prediction & Accuracy of $\hat y^{\sf aux}$    & MAE of $\hat y$ (lbs) \\
\midrule
LSTM \cite{e1}     & Concatenation & Yes & 0.874 ($\pm$0.033)           & 8.752 ($\pm$0.332)    \\
TCN \cite{m3}      & Concatenation & Yes & 0.974 ($\pm$0.002)           & 7.593 ($\pm$0.176)    \\ 
Transformer \cite{m6}      & Concatenation & Yes & 0.949 ($\pm$0.008)                     &      7.440 ($\pm$0.217)    \\ 
Informer \cite{e2}& Concatenation & Yes  & 0.964 ($\pm$0.008)           & 7.434 ($\pm$0.380)    \\ 
TimesNet \cite{e3} & Concatenation & Yes & 0.948 ($\pm$0.013)           & 8.040 ($\pm$0.189)   \\ 
\textbf{\texttt{AuxVAE} (ours)}      & \textbf{Cross Attention} & \textbf{Yes} & \textbf{0.982 ($\pm$0.007)} & \textbf{5.670 ($\pm$0.185)} \\
\bottomrule
\end{tabular}
}
\end{table}

\subsubsection{Experiment II: Ablation study}

In this section, we present a systematic ablation study to evaluate the effectiveness of each component in our \texttt{AuxVAE} model. Specifically, we examine how the use of auxiliary input ($\bX^{\sf aux}$) and output ($y^{\sf aux}$) and different auxiliary fusion strategies affect overall model performance.

Table \ref{tab:ablation} presents the results of our ablation study. The columns indicate whether the model incorporates auxiliary input (column `$\bX^{\sf aux}$') and/or auxiliary output supervision (column `$y^{\sf aux}$'), as well as the fusion strategy employed for the auxiliary input; either simple concatenation or the proposed bi-directional cross-attention mechanism (column ‘Fusion Type’). 

Without auxiliary data (Setting 1), the model predicts load magnitude solely based on the loaded gait, resulting in the highest regression error. When joint classification is introduced (Setting 2), the performance improves slightly, indicating that multi-task supervision can indirectly regularize the regression objective. Moreover, adding auxiliary gait via early feature concatenation (Setting 3) leads to further improvement in both classification accuracy and load prediction, demonstrating the value of personalized baseline motions in hand load estimation. However, the performance gain is limited, and classification accuracy unexpectedly drops slightly compared to Setting 2. This suggests that a naive concatenation may not effectively capture the complex temporal relationships between loaded and unloaded gait motions. Even in the absence of joint modeling (Setting 4), incorporating cross-attention significantly improves load estimation relative to naive input concatenation. Performance further improves when our proposed attention-based fusion is combined with joint modeling of auxiliary outputs (Setting 5), yielding the best overall results. These findings indicate that joint modeling of auxiliary outputs complements the structured fusion of auxiliary inputs, enabling the model to better capture complex temporal correspondences and thereby enhance predictive performance.

\begin{table}[htb!]
\centering
\caption{Ablation study on the integration of auxiliary baseline gaits and carrying styles.}
\label{tab:ablation}
\begin{tabular}{@{}cccc|cc@{}}
\toprule
Model         & $\bX^{\sf aux}$ & Fusion Type     & $y^{\sf aux}$  & Accuracy of $\hat y^{\sf aux}$ & MAE of $\hat y$ (lbs)     \\ \midrule
Ablation Setting 1    &  \xmark         & -               & \xmark         & -                         & 8.036 ($\pm$ 0.292) \\
Ablation Setting 2    &  \xmark         & -               & \cmark         & 0.957 ($\pm$ 0.010)       & 7.896 ($\pm$ 0.221) \\
Ablation Setting 3    &  \cmark         & Concatenation   & \cmark         & 0.955 ($\pm$ 0.012)       & 7.130 ($\pm$ 0.214) \\
Ablation Setting 4    &  \cmark         & Cross attention & \xmark         & -                         & 6.746 ($\pm$ 0.446) \\
\texttt{AuxVAE} (ours) &  \cmark         & Cross attention & \cmark         & \textbf{0.982} ($\pm$ 0.007) & \textbf{5.670} ($\pm$ 0.185) \\
\bottomrule
\end{tabular}
\end{table}

\subsection{Analysis and Discussion}

\subsubsection{Latent space visualization}

\begin{figure*}[htb!]
\centerline{\includegraphics[width=0.9\textwidth]{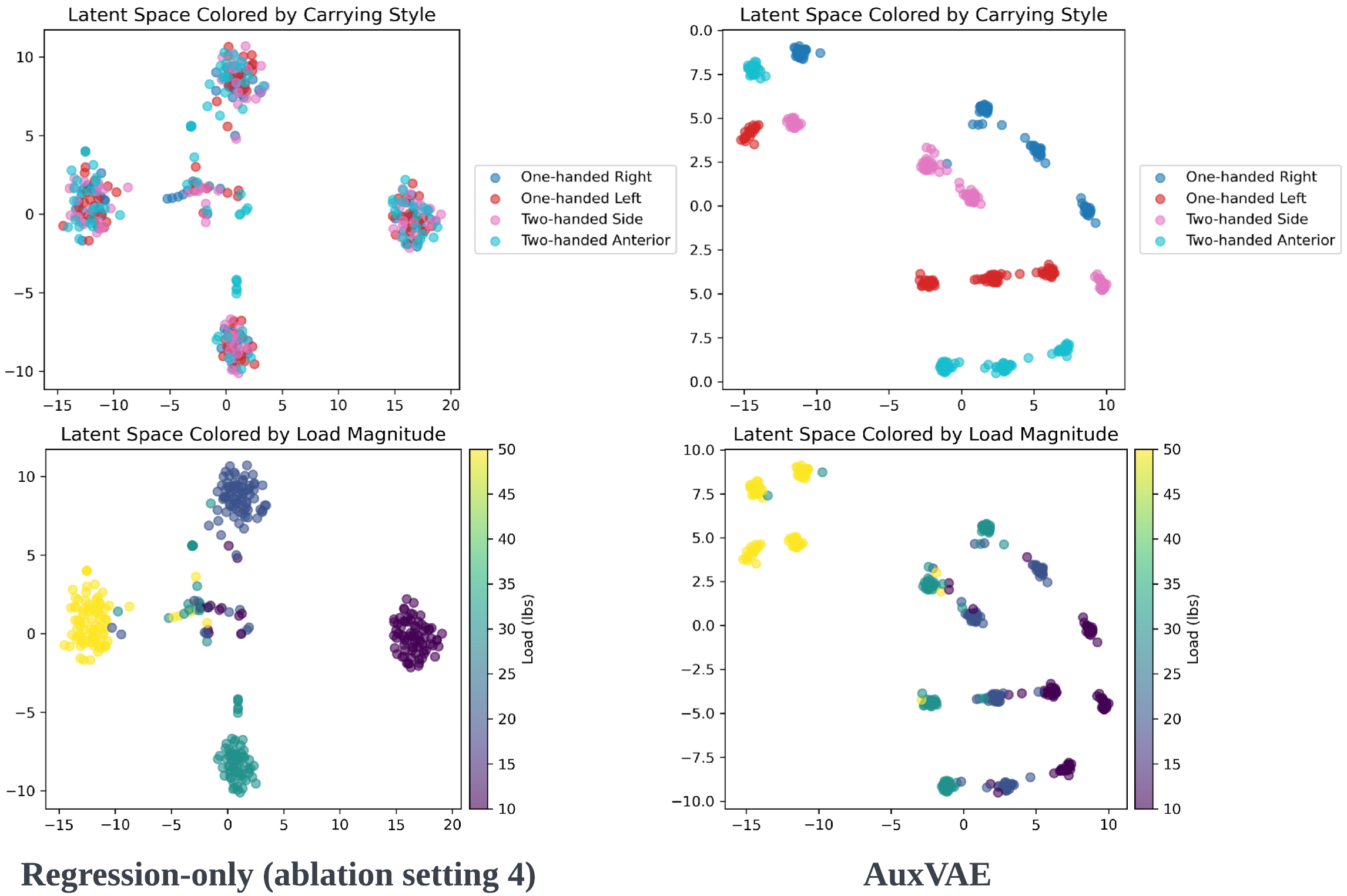}}
\caption{Latent space visualization. Top: colored by carrying style. Bottom: colored by load magnitude. Left: regression-only model. Right: joint model (\texttt{AuxVAE}).}
\label{fig:latent}
\end{figure*}

To further investigate how our joint modeling for primary and auxiliary outputs improves predictive performance, we visualize the learned latent spaces using t-SNE in Figure~\ref{fig:latent}. The panels on the left and right column visualize the spread of latent representations from the model trained only for regression (i.e., predicting load magnitude) and for both regression and classification (i.e., classifying carrying style) through our joint modeling, respectively. 

In the top two panels, latent representations are color-coded by carrying style. The latent representation clusters in the regression-only model exhibit a significant overlap across styles, suggesting that the latent space does not capture style-specific gait variations. In contrast, the joint model shows clearly separated clusters in terms of carrying style. This result indicates that our approach that accounts for carrying style encourages disentanglement across different carrying styles and enables the encoder to leverage auxiliary information on carrying style more explicitly.

In the bottom two panels, latent representations are color-coded by load magnitude. Although the regression-only model produces clusters by load magnitude, the latent space exhibits no clear trend with respect to weight levels. In contrast, intriguingly, the latent space learned through our joint modeling displays a more organized structure: latent representations align along a nearly linear trajectory, with heavier loads positioned in the upper left and progressively lighter loads extending toward the bottom right. This arrangement suggests that the model has captured a monotonic manifold in the latent space, where load magnitude is consistently encoded along a specific directional axis. It would be very interesting to explore the model’s ability to capture meaningful intrinsic physical characteristics of gait motion within the latent space, which could support the development of an extrapolative framework capable of generalizing to unseen load conditions beyond the training range.

\subsubsection{Auxiliary fusion via bidirectional cross-attention}

\begin{figure*}[htb!]
\centerline{\includegraphics[width=0.95\textwidth]{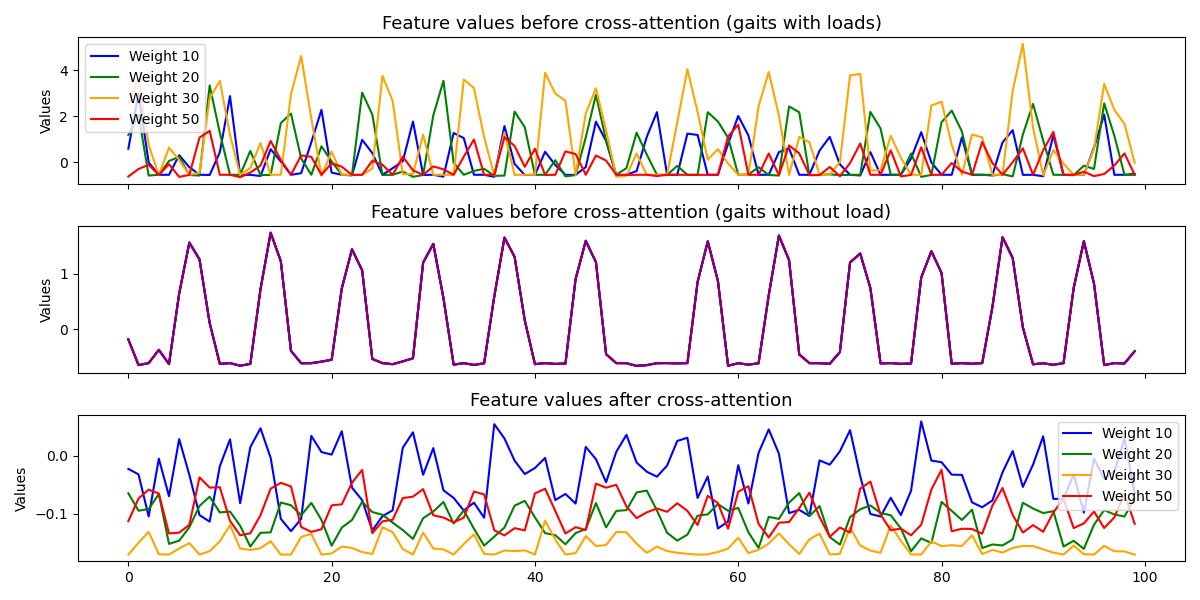}}
\caption{Feature trajectories before and after cross-attention across different load conditions. Top: raw loaded features. Middle: auxiliary gait features. Bottom: fused features post attention.}
\label{fig:attention}
\end{figure*}

Figure~\ref{fig:attention} illustrates the evolution of features across load magnitudes before and after the bidirectional cross-attention layer. We visualize the transformed representations of the original gait motion data at intermediate layers of our proposed model. Specifically, the top and middle panels show the transformed features with different hand loads and without hand load, respectively, prior to the cross-attention layer. The bottom panel presents the transformed features after auxiliary fusion via the bidirectional cross-attention mechanism.

As shown in the figure, the transformed features corresponding to hand-loaded gait data exhibit high variability and irregular temporal dynamics, with substantial fluctuations and inconsistent patterns across weight conditions. This indicates that predicting load magnitude based solely on loaded gait motion is non-trivial. In contrast, the proposed auxiliary fusion mechanism, leveraging baseline gait information through cross-attention, effectively enhances discriminative structure in the feature space. This fusion highlights load-specific patterns, resulting in more consistent representations and improved predictive performance.

\subsubsection{Is adding auxiliary information always helpful?}

One of the key findings of this study is the importance of incorporating auxiliary information through an explicit fusion mechanism to enhance predictive performance in gait-based hand load estimation. We hypothesize that naively introducing auxiliary information into the model input is unlikely to be effective, as discussed in Section \ref{sec:exp_results}. To further investigate this, we evaluated baseline models that incorporated auxiliary inputs $\bX^{\sf aux}$ and targets $y^{\sf aux}$ using a simple yet naive approach: concatenating $\bX^{\sf aux}$ with the primary input $\bX$, and jointly predicting both $y$ and $y^{\sf aux}$ using separate output layers appended to the final shared layers. Results are presented in Table \ref{tab:baseline}.

\begin{table}[htb!]
\centering
\caption{Performance of baseline models with simple auxiliary fusion.}
\label{tab:baseline}
\resizebox{0.35\textwidth}{!}{%
\begin{tabular}{@{}ccccc@{}}
\toprule
Model                  & MAE of $\hat y$ (lbs)  \\
\midrule
LSTM \cite{e1}         & 8.336 ($\pm$0.410)    \\
TCN \cite{m3}          & 6.613 ($\pm$0.184)    \\ 
Transformer \cite{m6}   & 9.316 ($\pm$0.715)    \\ 
Informer \cite{e2}     & 7.461 ($\pm$0.351)    \\ 
TimesNet \cite{e3}     & 8.307 ($\pm$1.070)   \\ 
\bottomrule
\end{tabular}
}
\end{table}

A comparison between Table~\ref{tab:baseline_comparison} and Table~\ref{tab:baseline} reveals that baseline models exhibit limited, and in some cases even negative, effects from the inclusion of auxiliary information and naive joint modeling. For instance, TCN performance deteriorates significantly, with MAE increasing from 6.613 to 7.593 (a 14.8\% decline), and LSTM shows a modest degradation from 8.336 to 8.752 MAE (5.0\% worse). Although transformer-based models show some gains—Transformer improves from 9.316 to 7.440 MAE (20.1\% better), Informer from 7.461 to 7.434 MAE (0.4\% better), and TimesNet from 8.307 to 8.040 MAE (3.2\% better); most improvements are marginal. These mixed or negative results suggest that simply concatenating auxiliary information and introducing joint prediction objectives does not reliably enhance performance across model architectures.

These findings yield two key insights. First, fusion through naïve concatenation may lead to \textit{information interference} rather than enhancement, particularly in the absence of a dedicated fusion mechanism. For example, when TCN processes concatenated loaded and unloaded gait signals, it appears to struggle with the increased dimensionality and potentially conflicting temporal patterns. Although unloaded gait provides a useful baseline, directly concatenating it with loaded patterns introduces confusion into the feature space, as the model lacks a mechanism to selectively emphasize relevant auxiliary signals. This interference is especially evident in architectures like TCN and LSTM, which process all input features uniformly and lack attention-based or adaptive weighting mechanisms.

Second, joint prediction introduces \textit{optimization complexity} that is not well supported by the baseline models. Simultaneously predicting load magnitude and carrying style imposes a multi-objective learning challenge, which may cause conflicts during training. For example, improving accuracy on carrying style classification could inadvertently degrade load prediction performance. This is reflected in the observed performance drop for several models. In contrast, our proposed joint modeling framework explicitly incorporates domain knowledge from ergonomics, recognizing that hand load estimation should be conditioned on carrying style \cite{r15}. This targeted design enables the model to exploit the auxiliary information more effectively, leading to improved predictive accuracy.

\section{Conclusion} \label{Conclusion}

Recent ML-based approaches for estimating biomechanical load exposure during manual material handling have shown substantial promise, particularly with the advancement of gait monitoring technologies using wearable sensors. However, most existing methods rely on learning a direct mapping from sensor-based motion data to hand load, leaving substantial room for performance improvement. This study addresses that gap by proposing a deep latent variable model equipped with explicit mechanisms to incorporate baseline gait and carrying style information, grounded in domain knowledge from ergonomics. Experimental results using a real-world dataset show the effectiveness of our proposed approach and highlight the need for our proposed auxiliary fusion mechanism over naive fusion methods.

We conclude the paper by outlining several promising directions to further enhance the proposed framework. First, given the sensitive nature of gait monitoring data, future work could explore privacy-preserving approaches that safeguard sensor data collected from individual users. In particular, integrating differential privacy techniques \cite{c1} or applying federated learning methods tailored for multi-sensor systems \cite{c2} would be valuable avenues to pursue. Second, due to the substantial variability in gait patterns across individuals, developing effective personalization strategies remains a critical challenge. Approaches such as few-shot learning \cite{c3} or real-time model adaptation \cite{c4} could enable rapid personalization of the hand load estimation model, allowing it to adapt to new users with only a small number of real-time sensor observations collected during deployment. Pursuing these directions would enable more practical and personalized applications of our approach in real-world ergonomic exposure assessments and diverse occupational environments.

\section*{Acknowledgment}

This research is supported by the National Science Foundation (Grant No. 2427599) and the National Safety Council through the Research to Solutions (R2S) Grant awards.

\bibliographystyle{ieeetr}
\bibliography{literature}

\begin{thebibliography}{10}

\bibitem{a1}
M.~Spallek, W.~Kuhn, S.~Uibel, A.~van Mark, and D.~Quarcoo, ``Work-related musculoskeletal disorders in the automotive industry due to repetitive work-implications for rehabilitation,'' {\em Journal of Occupational Medicine and Toxicology}, vol.~5, pp.~1--6, 2010.

\bibitem{a2}
Y.-C. Lee, X.~Hong, and S.~S. Man, ``Prevalence and associated factors of work-related musculoskeletal disorders symptoms among construction workers: a cross-sectional study in south china,'' {\em International Journal of Environmental Research and Public Health}, vol.~20, no.~5, p.~4653, 2023.

\bibitem{a3}
L.~Punnett and D.~H. Wegman, ``Work-related musculoskeletal disorders: the epidemiologic evidence and the debate,'' {\em Journal of electromyography and kinesiology}, vol.~14, no.~1, pp.~13--23, 2004.

\bibitem{a4}
M.~MassirisFern{\'a}ndez, J.~{\'A}. Fern{\'a}ndez, J.~M. Bajo, and C.~A. Delrieux, ``Ergonomic risk assessment based on computer vision and machine learning,'' {\em Computers \& Industrial Engineering}, vol.~149, p.~106816, 2020.

\bibitem{a5}
A.~Ranavolo, F.~Draicchio, T.~Varrecchia, A.~Silvetti, and S.~Iavicoli, ``Wearable monitoring devices for biomechanical risk assessment at work: Current status and future challenges—a systematic review,'' {\em International journal of environmental research and public health}, vol.~15, no.~9, p.~2001, 2018.

\bibitem{a6}
H.-H. Wang, W.-C. Tsai, C.-Y. Chang, M.-H. Hung, J.-H. Tu, T.~Wu, and C.-H. Chen, ``Effect of load carriage lifestyle on kinematics and kinetics of gait,'' {\em Applied bionics and biomechanics}, vol.~2023, no.~1, p.~8022635, 2023.

\bibitem{a7}
C.~Z.-H. Ma, Z.~Li, and C.~He, ``Advances in biomechanics-based motion analysis,'' 2023.

\bibitem{a8}
S.~Anwer, H.~Li, M.~F. Antwi-Afari, W.~Umer, I.~Mehmood, and A.~Y.~L. Wong, ``Effects of load carrying techniques on gait parameters, dynamic balance, and physiological parameters during a manual material handling task,'' {\em Engineering, Construction and Architectural Management}, vol.~29, no.~9, pp.~3415--3438, 2022.

\bibitem{a9}
J.~Yoon, B.~Lee, J.~Chun, B.~Son, and H.~Kim, ``Investigation of the relationship between ironworker’s gait stability and different types of load carrying using wearable sensors,'' {\em Advanced Engineering Informatics}, vol.~51, p.~101521, 2022.

\bibitem{a10}
S.~Lim and C.~D’Souza, ``Measuring effects of two-handed side and anterior load carriage on thoracic-pelvic coordination using wearable gyroscopes,'' {\em Sensors}, vol.~20, no.~18, p.~5206, 2020.

\bibitem{r1}
L.~McAtamney and E.~N. Corlett, ``Rula: a survey method for the investigation of work-related upper limb disorders,'' {\em Applied ergonomics}, vol.~24, no.~2, pp.~91--99, 1993.

\bibitem{r2}
S.~Hignett and L.~McAtamney, ``Rapid entire body assessment (reba),'' {\em Applied ergonomics}, vol.~31, no.~2, pp.~201--205, 2000.

\bibitem{r3}
Z.~Jiao, K.~Huang, Q.~Wang, G.~Jia, Z.~Zhong, and Y.~Cai, ``Improved reba: deep learning based rapid entire body risk assessment for prevention of musculoskeletal disorders,'' {\em Ergonomics}, vol.~67, no.~10, pp.~1356--1370, 2024.

\bibitem{r4}
M.~MassirisFern{\'a}ndez, J.~{\'A}. Fern{\'a}ndez, J.~M. Bajo, and C.~A. Delrieux, ``Ergonomic risk assessment based on computer vision and machine learning,'' {\em Computers \& Industrial Engineering}, vol.~149, p.~106816, 2020.

\bibitem{r5}
V.~Figueira, S.~Silva, I.~Costa, B.~Campos, J.~Salgado, L.~Pinho, M.~Freitas, P.~Carvalho, J.~Marques, and F.~Pinho, ``Wearables for monitoring and postural feedback in the work context: a scoping review,'' {\em Sensors}, vol.~24, no.~4, p.~1341, 2024.

\bibitem{r6}
M.~L. Nunes, D.~Folgado, C.~Fuj{\~a}o, L.~Silva, J.~Rodrigues, P.~Matias, M.~Barandas, A.~V. Carreiro, S.~Madeira, and H.~Gamboa, ``Posture risk assessment in an automotive assembly line using inertial sensors,'' {\em IEEE Access}, vol.~10, pp.~83221--83235, 2022.

\bibitem{r7}
P.~Giannini, G.~Bassani, C.~A. Avizzano, and A.~Filippeschi, ``Wearable sensor network for biomechanical overload assessment in manual material handling,'' {\em Sensors}, vol.~20, no.~14, p.~3877, 2020.

\bibitem{r8}
T.~Chatzis, D.~Konstantinidis, and K.~Dimitropoulos, ``Automatic ergonomic risk assessment using a variational deep network architecture,'' {\em Sensors}, vol.~22, no.~16, p.~6051, 2022.

\bibitem{r9}
C.~Zhou, J.~Zeng, L.~Qiu, S.~Wang, P.~Liu, and J.~Pan, ``An attention-based adaptive spatial--temporal graph convolutional network for long-video ergonomic risk assessment,'' {\em Engineering Applications of Artificial Intelligence}, vol.~131, p.~107780, 2024.

\bibitem{r10}
M.~F. Antwi-Afari, Y.~Qarout, R.~Herzallah, S.~Anwer, W.~Umer, Y.~Zhang, and P.~Manu, ``Deep learning-based networks for automated recognition and classification of awkward working postures in construction using wearable insole sensor data,'' {\em Automation in construction}, vol.~136, p.~104181, 2022.

\bibitem{r11}
M.~Trkov, D.~T. Stevenson, and A.~S. Merryweather, ``Classifying hazardous movements and loads during manual materials handling using accelerometers and instrumented insoles,'' {\em Applied ergonomics}, vol.~101, p.~103693, 2022.

\bibitem{r12}
L.~Li, S.~Prabhu, Z.~Xie, H.~Wang, L.~Lu, and X.~Xu, ``Lifting posture prediction with generative models for improving occupational safety,'' {\em IEEE Transactions on Human-Machine Systems}, vol.~51, no.~5, pp.~494--503, 2021.

\bibitem{r13}
I.~Conforti, I.~Mileti, Z.~Del~Prete, and E.~Palermo, ``Measuring biomechanical risk in lifting load tasks through wearable system and machine-learning approach,'' {\em Sensors}, vol.~20, no.~6, p.~1557, 2020.

\bibitem{r14}
E.~S. Matijevich, P.~Volgyesi, and K.~E. Zelik, ``A promising wearable solution for the practical and accurate monitoring of low back loading in manual material handling,'' {\em Sensors}, vol.~21, no.~2, p.~340, 2021.

\bibitem{r15}
S.~Lim and C.~D'Souza, ``Statistical prediction of load carriage mode and magnitude from inertial sensor derived gait kinematics,'' {\em Applied ergonomics}, vol.~76, pp.~1--11, 2019.

\bibitem{r16}
A.~Rahman, S.~Lim, and S.~Chung, ``Fairness in machine learning-based hand load estimation: A case study on load carriage tasks,'' {\em arXiv preprint arXiv:2504.05610}, 2025.

\bibitem{r17}
Y.~Cui and Y.~Kang, ``Multi-modal gait recognition via effective spatial-temporal feature fusion,'' in {\em Proceedings of the IEEE/CVF Conference on Computer Vision and Pattern Recognition}, pp.~17949--17957, 2023.

\bibitem{r18}
S.~Zou, J.~Xiong, C.~Fan, C.~Shen, S.~Yu, and J.~Tang, ``A multi-stage adaptive feature fusion neural network for multimodal gait recognition,'' {\em IEEE Transactions on Biometrics, Behavior, and Identity Science}, 2024.

\bibitem{r19}
S.~T.~Y. Aung and W.~Kusakunniran, ``A comprehensive review of gait analysis using deep learning approaches in criminal investigation,'' {\em PeerJ Computer Science}, vol.~10, p.~e2456, 2024.

\bibitem{m1}
D.~P. Kingma, M.~Welling, {\em et~al.}, ``Auto-encoding variational bayes,'' 2013.

\bibitem{m2}
K.~Sohn, H.~Lee, and X.~Yan, ``Learning structured output representation using deep conditional generative models,'' {\em Advances in neural information processing systems}, vol.~28, 2015.

\bibitem{m3}
S.~Bai, J.~Z. Kolter, and V.~Koltun, ``An empirical evaluation of generic convolutional and recurrent networks for sequence modeling,'' {\em arXiv preprint arXiv:1803.01271}, 2018.

\bibitem{m4}
H.~Tan and M.~Bansal, ``Lxmert: Learning cross-modality encoder representations from transformers,'' {\em arXiv preprint arXiv:1908.07490}, 2019.

\bibitem{m5}
D.~Hendrycks and K.~Gimpel, ``Gaussian error linear units (gelus),'' {\em arXiv preprint arXiv:1606.08415}, 2016.

\bibitem{m6}
A.~Vaswani, N.~Shazeer, N.~Parmar, J.~Uszkoreit, L.~Jones, A.~N. Gomez, {\L}.~Kaiser, and I.~Polosukhin, ``Attention is all you need,'' {\em Advances in neural information processing systems}, vol.~30, 2017.

\bibitem{m7}
C.~K. S{\o}nderby, T.~Raiko, L.~Maal{\o}e, S.~K. S{\o}nderby, and O.~Winther, ``Ladder variational autoencoders,'' {\em Advances in neural information processing systems}, vol.~29, 2016.

\bibitem{m8}
S.~R. Bowman, L.~Vilnis, O.~Vinyals, A.~M. Dai, R.~Jozefowicz, and S.~Bengio, ``Generating sentences from a continuous space,'' {\em arXiv preprint arXiv:1511.06349}, 2015.

\bibitem{e1}
A.~Graves and A.~Graves, ``Long short-term memory,'' {\em Supervised sequence labelling with recurrent neural networks}, pp.~37--45, 2012.

\bibitem{e2}
H.~Zhou, S.~Zhang, J.~Peng, S.~Zhang, J.~Li, H.~Xiong, and W.~Zhang, ``Informer: Beyond efficient transformer for long sequence time-series forecasting,'' in {\em Proceedings of the AAAI conference on artificial intelligence}, vol.~35, pp.~11106--11115, 2021.

\bibitem{e3}
H.~Wu, T.~Hu, Y.~Liu, H.~Zhou, J.~Wang, and M.~Long, ``Timesnet: Temporal 2d-variation modeling for general time series analysis,'' {\em arXiv preprint arXiv:2210.02186}, 2022.

\bibitem{e4}
Y.~Wang, H.~Wu, J.~Dong, Y.~Liu, M.~Long, and J.~Wang, ``Deep time series models: A comprehensive survey and benchmark,'' 2024.

\bibitem{c1}
M.~Abadi, A.~Chu, I.~Goodfellow, H.~B. McMahan, I.~Mironov, K.~Talwar, and L.~Zhang, ``Deep learning with differential privacy,'' in {\em Proceedings of the 2016 ACM SIGSAC conference on computer and communications security}, pp.~308--318, 2016.

\bibitem{c2}
J.~Gao and S.~Chung, ``Federated automatic latent variable selection in multi-output gaussian processes,'' {\em arXiv preprint arXiv:2407.16935}, 2024.

\bibitem{c3}
C.~Finn, P.~Abbeel, and S.~Levine, ``Model-agnostic meta-learning for fast adaptation of deep networks,'' in {\em International conference on machine learning}, pp.~1126--1135, PMLR, 2017.

\bibitem{c4}
S.~Chung and R.~Al~Kontar, ``Real-time adaptation for time-series signal prediction using label-aware neural processes,'' {\em Reliability Engineering \& System Safety}, p.~110833, 2025.

\end{thebibliography}

\end{document}